# Generative Artificial Intelligence in Healthcare: Ethical Considerations and Assessment Checklist

Yilin Ning[1#], Salinelat Teixayavong[1#], Yuqing Shang[1], Julian Savulescu[2,3], Vaishaanth Nagaraj[4], Di Miao[1], Mayli Mertens[5,6], Daniel Shu Wei Ting[1,7,8], Jasmine Chiat Ling Ong[9], Mingxuan Liu[1], Jiuwen Cao[10,11], Michael Dunn[2], Roger Vaughan[1,12], Marcus Eng Hock Ong[12,13], Joseph Jao-Yiu Sung[14,15], Eric J Topol[16], Nan Liu[1,12,17]*

[1]Centre for Quantitative Medicine, Duke-NUS Medical School, Singapore, Singapore
[2]Centre for Biomedical Ethics, National University of Singapore, Singapore, Singapore
[3]Wellcome Centre for Ethics and Humanities, University of Oxford, Oxford, UK
[4]School of Medicine, Imperial College London, London, UK
[5]Centre for Ethics, Department of Philosophy, University of Antwerp, Antwerp, Belgium
[6]Antwerp Center on Responsible AI, University of Antwerp, Antwerp, Belgium
[7]Singapore Eye Research Institute, Singapore National Eye Centre, Singapore, Singapore
[8]SingHealth AI Office, Singapore Health Services, Singapore, Singapore
[9]Division of Pharmacy, Singapore General Hospital, Singapore, Singapore
[10]Machine Learning and I-Health International Cooperation Base of Zhejiang Province, Hangzhou Dianzi University, Zhejiang, China
[11]Artificial Intelligence Institute, Hangzhou Dianzi University, Zhejiang, China
[12]Programme in Health Services and Systems Research, Duke-NUS Medical School, Singapore, Singapore
[13]Department of Emergency Medicine, Singapore General Hospital, Singapore, Singapore
[14]Lee Kong Chian School of Medicine, Nanyang Technological University, Singapore, Singapore
[15]State Key Laboratory of Digestive Disease, The Chinese University of Hong Kong, Hong Kong SAR, China
[16]Scripps Research Translational Institute, Scripps Research, La Jolla, CA, USA
[17]Institute of Data Science, National University of Singapore, Singapore, Singapore

[#]Equal contribution
*Correspondence: Nan Liu, Centre for Quantitative Medicine, Duke-NUS Medical School, 8 College Road, Singapore 169857, Singapore
Email: liu.nan@duke-nus.edu.sg

## Abstract

The widespread use of ChatGPT and other emerging technology powered by generative artificial intelligence (GenAI) has drawn much attention to potential ethical issues, especially in high-stakes applications such as healthcare, but ethical discussions are yet to translate into operationalisable solutions. Furthermore, ongoing ethical discussions often neglect other types of GenAI that have been used to synthesise data (e.g., images) for research and practical purposes, which resolved some ethical issues and exposed others. We conduct a scoping review of ethical discussions on GenAI in healthcare to comprehensively analyse gaps in the current research, and further propose to reduce the gaps by developing a checklist for comprehensive assessment and transparent documentation of ethical discussions in GenAI research. The checklist can be readily integrated into the current peer review and publication system to enhance GenAI research, and may be used for ethics-related disclosures for GenAI-powered products, healthcare applications of such products and beyond.

## Introduction

The recent developments of ChatGPT[1] and other large language model (LLM)-powered chatbots have drawn the attention of the general public, researchers and stakeholders to the fast-developing technology of generative artificial intelligence (GenAI). As the name suggests, the capability of GenAI to generate realistic content differentiates it from other general AI technology. OpenAI's ChatGPT and other LLM-powered chatbots (e.g., Google's Bard[2]) focus on text-based interactions with users and are actively extending to images and beyond, whereas tools like Midjourney[3] and OpenAI's DALL-E[4] focus on generating images based on user instructions. Although vastly enabling, the use of such technology has raised many ethical concerns including, but not limited to, biased and misleading information, privacy breaches and copyright violation.[5–7] Hence, there have been active discussions on and urgent calls for new guidelines, regulations and legislations on GenAI,[8–10] with some specifically for high-stakes applications such as in healthcare and education.[11–13]

Emerging GenAI-powered technology has been quickly rolled out to a huge and diverse user community, and this wide impact explains the enthusiastic discussions on its ethical implications in social media and the research community. A lot of discussions focus on ethical issues arising from the use of this emerging technology, e.g., ChatGPT was banned in Italy for a brief period to address and clarify privacy concerns.[14,15] On the other hand, other types of GenAI are less often in the centre of ethical discussions. As an example, the generative adversarial network (GAN), proposed in 2014,[16] is a typical type of GenAI method that has been widely used to generate images (and other types of data) for various purposes, e.g., to enhance medical images for improved diagnosis,[17] to mask personal information from images and videos for privacy protection,[18,19] and to create synthetic medical data for research use instead of the highly sensitive original data.[20–22] Indeed, GAN (and related GenAI methods[23]) have been used as solutions to some ethical concerns in healthcare applications, especially for preserving data privacy. A closer investigation of existing discussions on the role of various types of GenAI in causing and mitigating ethical concerns would provide a more holistic view on this important topic, and ultimately help identify actionable points for GenAI developers and practitioners. In this work, we systematically study the relevant literature in healthcare in a scoping review to understand the responses of the research community to potential ethical issues of GenAI research in healthcare, and use the findings to inform the operationalisation of existing ethical guides and recommendations by developing a checklist to facilitate responsible development and use of GenAI in healthcare and beyond.

## Methods

Our systematic scoping review on ethical discussions associated with GenAI in healthcare followed the Preferred Reporting Items for Systematic Reviews and Meta-Analyses extension for scoping reviews (PRISMA-ScR) guideline.[24]

**Search strategy and selection criteria**

We searched for articles with a set of search terms associated with three main concepts, "AI ethics", "Generative AI" and "Healthcare", published from January 2013 to July 2023 in four major academic research databases, i.e., PubMed, EMBASE (Excerpta Medica Database, Ovid), Web of Science and Scopus. Detailed definitions of the three concepts, search terms and search strategy are described in Supplementary eMethods A and Supplementary Table S1.

We excluded articles that were not within the domain of healthcare, did not apply GenAI, or did not discuss ethical issues in relation to the use of GenAI in healthcare. We also excluded articles that were not peer reviewed, not published as research articles (e.g., conference posters, conference abstracts, or book chapters), not full-length (i.e., articles citing no more than 10 references), not written in English. A main aim of this scoping review is to understand the current ethical assessments in GenAI research and to inform future practice, therefore we restricted our review to the published scientific literature that reflects a reasonable quality of reporting standard for ethical considerations that is accepted by the research community, and did not include the grey literature (e.g., preprints).

**Data analysis**

Articles found were first divided into three portions and screened based on title and abstract by three pairs of independent reviewers (S.T. and V.N.; Y.N. and S.Y.; Y.N. and D.M.) following the exclusion criteria described above. Uncertainties and conflicts were resolved via discussion with S.T. and Y.N.. Articles included were further screened based on the full text using the same set of exclusion criteria.

From included articles we extracted information on seven variables pertaining article type, GenAI and ethics: (1) whether the article describes original research (i.e., development or application of GenAI methods) or review-type work (e.g., review, viewpoint, or editorial that specifically covered GenAI), (2) data modalities of GenAI application, (3) GenAI model(s) discussed, (4) role of GenAI (i.e., whether it caused or resolved ethical issues), (5) ethical issues discussed, (6) if GenAI caused ethical issue(s), whether any solution was proposed, and (7) whether the article had dedicated discussions on ethical issues (as opposed to a brief mention of ethics in background information or general discussion).

To better summarise the ethical issues discussed in the included articles, we categorised them through a codification system into nine overarching ethical principles that have been identified to be most pertinent across AI ethics guidelines and in application of AI within the healthcare settings.[10,25] Box 1 details the adopted definitions and Table 1 lists codes associated with each principle summarised based on aforementioned discussion on ethics of AI for healthcare. Likewise, we summarised data modality of GenAI application into three general categories, namely text, image (including video) and structured (e.g., tabular data, and signal data such as electrocardiogram or speech signal). Information on ethical principles discussed and the role of GenAI by data modality was analysed using evidence gap maps to understand the current research landscape in healthcare.

**Checklist development**

In a recent work that proposed an ethical framework of AI for healthcare for AI developers, the authors endorsed the need to work with health AI practioners to develop ethical AI checklists as a means to operationalise considerations and solutions to a list of ethical issues identified.[26] Hence, after analysing the articles identified in our scoping review and summarising prominent gaps in current GenAI research (which will be elaborated in the following sections), we used the findings to inform the development of a Transparent Reporting of Ethics for Generative Artificial Intelligence (TREGAI) checklist to promote systematic assessment and transparent reporting of ethical considerations in GenAI research. Specifically, based on observations from the scoping review (especially on a subset of articles with a stronger ethics focus), we selected a set of well-established ethical principles in the AI ethics literature to include in our checklist that are essential to and operationalisable in GenAI research in healthcare, and associated each principle with possible actions for operationalisation. Detail on checklist development, specifically the choice of ethical principles and distinction from existing guides on AI, is described in the Supplementary eMethods B. The structure, application settings and potential impact of the proposed TREGAI checklist will be elaborated in Discussion.

## Results

Our scoping review identified 1417 unique articles (see Supplementary Figure S1 for the PRISMA flow diagram), of which 193 articles were included for analysis. Detailed information extracted on these articles are presented in Supplementary Table S2. The 193 articles included were published between 2018 and 2023, of which 31 articles are review-type work and the other 162 articles are original research. Specifically, 20 review-type articles and nine original research articles were motivated by ethical considerations or had dedicated ethical discussions, whereas another 66 articles investigated ethics-related considerations from quantitative or methodological perspectives (see Supplementary Table S2).

While three articles discussed GenAI for more than one data modalities (with two articles covering text and image, and one article covering text, image, and structured data), most articles reviewed were dedicated to single data modality when discussing ethical issues. Our analysis of these articles revealed notable differences in how researchers approach ethical concerns by data modality, especially on the ethical concerns discussed and the role of GenAI in causing or resolving these concerns, which we summarised in an evidence gap map (see Figure 1).

**GenAI for text data-based healthcare**

Forty-one of the 193 articles discussed ethical considerations pertaining to GenAI applications for text data, with 20 articles describing methodological developments or applications of GenAI and the other 21 articles describing review-type works on this topic. Although some of these review-type articles used the general term “generative AI”, the main body and supporting evidence focused on LLMs. Twenty-eight articles investigated or had in-depth discussions on ethical issues, whereas the other 13 articles only briefly touched on some ethical aspects.

Among the 41 articles, 29 articles focused on discussing ethical issues caused by LLMs (and specifically by GPT in 16 of the articles), covering a wide range of application scenarios and considered the application of all nine ethical principles defined in Box 1 (see Figure 1), as well as other less discussed concerns (which we summarise in the "Others" category) such as moral aspects (e.g., compassion) of LLM outputs, human-AI interaction, and the rights of LLMs to be considered as co-authors in scientific papers. One paper only commented briefly on the need for ethical considerations in LLMs and is summarised in the "Others" category. Although all ethical principles are important, some are discussed more often than others, e.g., non-maleficence (also referred to in the literature as "benevolence"), equity, and privacy.

Fifteen of the 41 articles aimed to resolve some existing ethical issues (for example, confidentiality of medical data) by using LLMs and other GenAI (e.g., GAN, autoencoder or diffusion), such as, to reduce privacy concerns by generating synthetic medical text, to reduce disparity by providing accessible services and assistance, to detect health-related misinformation, to generate trusted content, and to improve accountability or transparency over existing approaches. While most articles focused on either identifying ethical issues caused by GenAI or proposing GenAI-based solutions, three articles discussed both to provide a more balanced perspective.

**GenAI for image and structured data-based healthcare**

Unlike the diverse application scenarios of GenAI based on text data, for image and structured data, this use of GenAI focuses on data synthesis and encryption. Hence most articles discussed the methodological developments of GenAI as giving rise to a more distinctive and focused set of ethical issues.

Notably, of the 98 articles on image data and 58 articles on structured data, more than half (n=63 for image data and n=33 for structured data) only mentioned ethical considerations as a brief motivation for methodological developments or as a general discussion point. The rest included more in-depth discussions or (mostly quantitative) evaluations of ethical issues. Among these 155 articles (as one article covered multiple modalities), 11 articles were review-type work, where ten articles reviewed methods that mentioned one or two ethical perspectives, and only one article[27] discussed detailed ethical concerns on GenAI applications.

Resolving privacy issues was the main aim of articles for these two data modalities (n=74 for image data and n=50 for structured data; see Figure 1), predominantly by generating synthetic data using GAN. Eight articles on image data and nine articles on structured data used GenAI to reduce bias, e.g., by synthesising data for under-represented subgroups in existing databases. For both data modalities, we did not see explicit discussions on resolving autonomy or integrity issues using GenAI, and for structured data the articles additionally lacked discussions on trust or transparency.

Only 11 articles for image data selectively discussed some ethical issues that GenAI can give rise to, without specific discussions regarding autonomy or integrity. For structured data, only four articles discussed equity, privacy, or data security issues caused by GenAI. Only two articles on structured data included both the cause and resolving perspectives by discussing ethical issues that may arise from limitations of methods proposed, specifically bias induced when synthesising data to resolve privacy issues.

## Discussion

Despite the rising number of articles discussing ethical concerns on GenAI in healthcare, some important aspects are lacking in the current literature. Our scoping review systematically summarises the inadequacies in the current literature, which we elaborate below as four gaps. Furthermore, as our response to the gaps identified, in this section we introduce the TREGAI checklist, and elaborate how it may contribute to more responsible GenAI research in healthcare and in broader application settings.

### Gaps in ethical discussions on GenAI

First, in current GenAI research there lack solutions for ethical issues. For articles focusing on identifying ethical issues caused by GenAI, regulations and guidelines were most frequently raised as a solution. This was observed from 11 of the 29 articles discussing ethical issues caused by LLMs and the only article that reviewed ethical issues caused by GenAI for image data. Among the other 18 articles on ethical issues caused by LLMs, 13 articles did not discuss any solutions, and other solutions mentioned included ensuring doctor autonomy over LLM output, improving user awareness of limitations of LLMs, and implementing security technology. While guidelines and regulations for appropriate use of GenAI are important, difficulties can arise when applying ethical guidance based on a set of principles. This is because it can be difficult to interpret the specific requirements of a broad ethical principle in any given context in which GenAI is applied, or because the guidance offers limited assistance when trade-offs need to be made between ethical principles.

Moreover, due to the complexity and fast advancements in methods and technology, compliance to well-established legal regulations does not necessarily prevent ethical breaches. As examples, the privacy rule enforced by the Health Insurance Portability and Accountability Act of the United States Congress is insufficient to prevent privacy breaches (and a few other ethical issues) by the advanced technology employed by LLMs,[28] and the more recent European AI Act is inadequate in aspects like formal AI definition and risk management.[8] Similarly, general users of GenAI are unlikely to understand the technology well enough to prescribe a reasonable level of trust on its output or identify potential misinformation. On the contrary, the non-human nature and the confident and professional tone of well-designed LLMs can earn unwarranted trust from lay users,[28,29] potentially leading to more privacy leaks and harms from (partially) incorrect or biased information.

Secondly, there is insufficient discussion on ethical concerns beyond LLMs. As highlighted in the previous section, most dedicated ethical discussions focus on LLMs, despite the use of other GenAI methods (e.g., GAN) for text data and more prominently for other data modalities. Indeed, LLM-powered chatbots such as ChatGPT make the powerful tool easily accessible to healthcare professionals, medical students, and the general public without much need for technical background, hence substantially increasing the impact of any resulting ethical issues. However, inadequacies or concerns associated with other GenAI methods can also affect healthcare directly, or indirectly in the long run. For example, GAN-based approaches have been exploited for insurance scams by editing or injecting fake medical images or targeted in cyberattacks to steal confidential information,[30,31] among other malicious activities, but such topics are more often discussed from purely technical rather than ethical perspectives. Synthetic data is proposed as a solution to medical data deficiency for research purposes due to privacy and security concerns,[20] for which GAN-based approaches are becoming (if have not already become) the state-of-the-art. Compared to the large body of research on data synthesising methods, less investigated are benchmarks to quantify the quality of synthetic data in preserving patient privacy, important characteristics of the original data and minority patient groups to support unbiased and trustworthy future research,[20–22,32,33] among other ethical gaps illustrated in Figure 1.

Thirdly, there lacks a common and comprehensive framework for ethical discussions. While viewpoints and perspectives in leading journals are discussing a wide range of ethical issues that may arise (or have arisen) from applications of LLMs in healthcare and other settings, articles on GenAI for image or structured data have been largely focusing on a restricted set of issues, particularly privacy, that may be directly resolved via methodological or technological developments. Ethical discussions can be challenging for GenAI researchers and developers as they are often not formally trained in ethics, but the inclusion of dedicated ethical discussions beyond main objectives in some recent works[34,35] shows an increasing interest of the research community in more in-depth discussions on such issues. Recommendations and guides have been made on the general application of AI for healthcare,[10,26,36,37] but currently they do not seem to be consistently operationalised in GenAI applications.

Next, different authors may endorse different definitions of ethical terms, or may select a subset of ethical keywords for discussion without clear justification. For example, the well-established ethical principle of "beneficence" that could be drawn on to motivate the use of GenAI on the grounds of enhancing the patient's well-being was lacking in the literature, which is a striking omission. While some ethical principles may be less relevant in some application scenarios than others (e.g., the use of GenAI to synthesise structured data may not have as direct impact on autonomy as for text data), such statements should be made explicitly by researchers (or developers) with reasonable justifications rather than be inferred post-hoc by interested users of the method. Incomplete ethical discussions may lead to insufficient methodological developments or questionable applications of existing GenAI methods.

Finally, there lack discussions on multimodal GenAI. As mentioned above, most articles reviewed involved unimodal GenAI, i.e., models that accept and generate data in a single modality. Among the three articles that involved multiple modalities, only two discussed multimodal GenAI (specifically GAN) that simultaneously generated chest X-Ray images and radiology reports, and the third article merely reviewed unimodal GANs for various modalities. Although multimodal GenAI is not yet widely applied in healthcare,[38] notable progress has been made in other fields, e.g., Google's visual language model extends LLMs to work with images,[39] and OpenAI is updating ChatGPT to handle voice and image input.[40] These extended LLMs will certainly stimulate useful applications in healthcare and beyond, but the increased complexity in the models and application settings will further complicate model evaluation for reliability (which is already challenging for existing LLMs[41–43]), and the extended capability and wide adoption of such technology can amplify the impact of related ethical issues.

Meanwhile, ongoing research is developing multimodal GenAI from a creative yet concerning approach: to reconstruct input to subjects (be it images or text) by analysing their functional magnetic resonance imaging (fMRI).[35,44–46] This might be called "reverse-mindreading". Each of these works focused on a single input modality, but it has been shown that GenAI trained on one modality can be applied to other modalities with minimal adjustments.[46] Although these works provided additional insights on brain functions and potentially healthcare benefits, the direct extraction of information from brain activities beyond health-related purposes poses important concerns in neuroethics.[47,48] Surprisingly, only two of the four papers explicitly discussed these kinds of ethical concerns, where one paper performed additional experiments to demonstrate preservation of patient privacy,[46] and the other paper only highlighted the general need for regulations.[35] A more disciplined approach is needed to ensure ethical use of medical data (including but not limited to fMRI) when developing multimodal GenAI for and beyond healthcare applications.

**Proposed checklist for ethical GenAI for healthcare**

To address gaps identified above require collaborative efforts. Despite the controversies on how to allocate responsibilities and credibility to harms and benefits caused by GenAI,[49] researchers (and healthcare professionals) who develop new GenAI (or modify existing ones) should be responsible for understanding and disclosing the capabilities and limitations of the tools developed or used, and those who apply existing GenAI in their research should be able to justify and discuss the appropriateness and potential issues of the tools used in-context. Some articles reviewed have started to call for actions from GenAI developers to reduce ethical issues, e.g., by highlighting the responsibilities of LLM developers in preventing ethical issues from arising in the first place,[12] and by developing benchmarks to evaluate the ethics of LLMs to facilitate future mitigation,[34] but such discussions are not easily integrated into future development, extension and application of GenAI without detailed actionable guides. Hence, we propose to reinforce ethical considerations in GenAI research in healthcare by mandating standardised and transparent reporting during peer review via our proposed TREGAI checklist (see Table 2).

Based on a development of the nine established ethical principles in Box 1 and the additional important principle of Beneficence, our suggested TREGAI checklist prompts researchers to possible actions to operationalise each consideration observed from the scoping review, e.g., identification of ethical issues caused or mitigation of existing issues, and if applicable, solutions for issues identified or inadequacies of mitigation methods. The checklist facilitates transparent and systematic documentation of ethical discussions by requiring researchers to check all types of discussions present in the manuscript relevant to each ethical principle and indicate corresponding text position for peer review, and can be appended with additional ethical considerations as relevant, e.g., when working with multimodal GenAI discussed above. When an ethical principle is deemed not applicable, researchers are strongly encouraged to justify this conclusion in the manuscript. They should also indicate in the checklist the involvement of philosophers or ethicists in the process of identifying potential issues, as other experts in their respective fields may miss important issues. Likewise, ethical discussions are preferably peer-reviewed by researchers with expertise in philosophy or ethics and reflected in the checklist. The use of the TREGAI checklist is demonstrated in a worked example in Supplementary eResults.

The TREGAI checklist we propose should not be considered as a replacement for serious, in-depth analysis of (potential) ethical concerns, as checklists have limitations[50] and "checklist ethics" is especially notorious for undermining thorough ethical reflection. On the contrary, it should be used to inform further investigations, preferably in collaboration with philosophers and ethicists both through our proposed peer-review process and, next, through institutional review board (IRB) review as required. As previously discussed,[51] healthcare professionals, AI experts and ethicists may have separately investigated and addressed overlapping ethical aspects in the respective literature (e.g., clinical AI fairness), which need to be combined for comprehensive understanding and improved solutions via effective cross-disciplinary collaboration. By accompanying each new research with a detailed ethics checklist during peer review, and possibly publishing the checklist as a supplementary document to the final publication, we at least strive for systematic ethical assessments of GenAI studies that facilitates more responsible and reliable applications. As GenAI is constantly evolving, we maintain the TREGAI checklist live online (see https://github.com/nliulab/GenAI-Ethics-Checklist) to allow timely updates, where we could incorporate additional ethical principles (e.g., from existing literature[26,36] and emerging multimodal GenAI research), updates in recommended actions, or future progress in GenAI regulations and guidelines to facilitate a disciplined, comprehensive and transparent reinforcement. As the TREGAI checklist focuses on ethical considerations of GenAI, research works developing GenAI could use it in addition to existing model development guidelines (e.g., TRIPOD[52,53] or CLAIM[54,55]) and other tools to enhance ethical discussion (e.g. the Ethical OS Toolkit[56]), as appropriate. Use of GenAI (e.g., ChatGPT) to assist the research process (e.g., to draft the manuscript) should refer to the upcoming CANGARU checklist for more specific guides.[57]

**Future perspectives**

With the growing application of AI for healthcare and other high-stakes fields, there is an expanding literature on ethical concerns and guidelines for applications of AI,[26,36,58,59] and similar discussions are extending to GenAI,[12,25,27] but they are not easily translated into improved research practice.[26] Our scoping review comprehensively analyses ethical discussions on GenAI in the context of healthcare to highlight the current lack of a systematic assessment of issues in all relevant application settings (e.g., across model types and data modalities) and corresponding solutions, and the reliance on regulations and governance to reinforce ethical standards. However, in view of the complexity of GenAI, it is critical to incorporate ethical considerations during the development and implementation phase, instead of post-hoc mitigations when issues arise. Hence, we highlight the accountability of researchers and developers to comprehensively discuss, fully disclose, and when feasible duly resolve, a well-defined set of relevant ethical issues in GenAI research, and develop the TREGAI checklist as a facilitating tool for scientific publication and fund application by providing a quick reference to relevant text in research articles or proposals. As an immediate extension, the TREGAI checklist can also be used to document ethical considerations detailed in user manuals of GenAI-powered products derived from research works for transparent disclosure to users. When some ethical issues are not yet fully resolvable, full disclosure of current limitations, including what have and have not been covered in current ethical assessments, allows interested users of GenAI methods and products to weigh possible harms against anticipated benefits in their intended application scenarios.

As explained in Supplementary eMethods B, our scoping review may not capture all relevant ethical issues, e.g., ethical considerations such as sustainability that is less discussed in GenAI research for healthcare. Nonetheless, our scoping review reasonably reflects the current imbalance and inadequacies in ethical discussions on GenAI for healthcare to inform future research. As an initial step towards more responsible GenAI research, our TREGAI checklist does not include some ethical considerations that are not easily operationalised (e.g., morality and dignity), but researchers can collaborate with ethicists to append additional ethical considerations to the checklist as appropriate. Moreover, as highlighted above, we strongly advocate GenAI researchers to collaborate with ethicists for more in-depth ethical investigations, and if applicable, modify the definitions of ethical principles to suit local jurisdictions, culture and application scenarios. While the TREGAI checklist targets research-related settings (e.g., publication and funding applications), it may be modified for GenAI-created contents (e.g., social media posts or teaching materials) to disclose a summary of benefits, limitations, and potential risks, where the social media platforms and educational institutions may take the responsibility to assemble a team of ethicists to advise on and review the disclosure of ethical issues. By maintaining the TREGAI checklist as a live document online (https://github.com/nliulab/GenAI-Ethics-Checklist), we keep it up-to-date to the fast development and expanding applications of GenAI.

## Conclusion

GenAI is a powerful technology with various potential roles in healthcare and beyond, and failing to meet ethical standards in these roles can have varying impact on daily life, e.g., from inefficient administrative services to worse health outcomes. It is critical to point out that it is early in the era of GenAI, especially with respect to its utility or any implementation in medical practice. The number of studies published until this scoping review is relatively limited, but identifying these issues that are essential features of GenAI as early as possible may ultimately help to promote trust and adoption for clinicians, patients and the general public. By suggesting the TREGAI checklist for ethical GenAI and maintaining it live to incorporate updated understanding and regulations, we advocate a systematic and balanced assessment of ethical considerations beyond standard methodological and technological perspectives, which may be extended to general AI to facilitate more responsible and trustworthy development of technology.

**Author contributions**

Y.N. and S.T. contributed equally. Initial development of ideas: Y.N., S.T., N.L. Acquisition, analysis, and interpretation of data: Y.N., S.T., Y.S., V.N., D.M. Drafting of the manuscript: Y.N. and S.T. Critical revision of the manuscript: Y.N., J.S., M.M., J.C.L.O., M.D., J.J.S., E.T., N.L. Interpretation of the content: all authors. Revisions of the manuscript: all authors. Final approval of the completed version: all authors. Overseeing the project: N.L.

**Declaration of interests**

This work was supported by the Duke-NUS Signature Research Programme funded by the Ministry of Health, Singapore. Any opinions, findings and conclusions or recommendations expressed in this material are those of the author(s) and do not reflect the views of the Ministry of Health. M.M. is funded by the European Union, through the HORIZON-MSCA-2022-PF-01-01 Marie Curie Postdoctoral Fellowship (Grant number 101107292 'PredicGenX'). This research was funded in whole, or in part, by the Wellcome Trust [Grant number WT203132/Z/16/Z]. For the purpose of open access, the author has applied a CC BY public copyright licence to any Author Accepted Manuscript version arising from this submission. Other authors declare no competing interests.

**Acknowledgement**

The funders of the study had no role in study design, data collection, data analysis, data interpretation, or writing of the report.

**Data sharing**

Search strategy and information extracted from articles reviewed are available in the Supplementary Material.
Correspondence and requests for materials should be addressed to Nan Liu.

**Reference**

1 OpenAI. ChatGPT. https://chat.openai.com (accessed Oct 2, 2023).

2 Bard – Chat-based AI tool from Google, powered by PaLM 2. https://bard.google.com (accessed Oct 2, 2023).

3 Midjourney. Midjourney. https://www.midjourney.com/home/ (accessed Oct 2, 2023).

4 DALL·E 2. https://openai.com/dall-e-2 (accessed Oct 2, 2023).

5 Health TLD. ChatGPT: friend or foe? *The Lancet Digital Health* 2023; **5**: e102.

6 Jones N. How to stop AI deepfakes from sinking society — and science. *Nature* 2023; **621**: 676–9.

7 Stokel-Walker C, Van Noorden R. What ChatGPT and generative AI mean for science. *Nature* 2023; **614**: 214–6.

8 Hacker P, Engel A, Mauer M. Regulating ChatGPT and other Large Generative AI Models. In: Proceedings of the 2023 ACM Conference on Fairness, Accountability, and Transparency. New York, NY, USA: Association for Computing Machinery, 2023: 1112–23.

9 EU AI Act: first regulation on artificial intelligence | News | European Parliament. 2023; published online Aug 6. https://www.europarl.europa.eu/news/en/headlines/society/20230601STO93804/eu-ai-act-first-regulation-on-artificial-intelligence (accessed Oct 2, 2023).

10 Jobin A, Ienca M, Vayena E. The global landscape of AI ethics guidelines. *Nat Mach Intell* 2019; **1**: 389–99.

11 Meskó B, Topol EJ. The imperative for regulatory oversight of large language models (or generative AI) in healthcare. *npj Digit Med* 2023; **6**: 1–6.

12 Abd-Alrazaq A, AlSaad R, Alhuwail D, *et al.* Large Language Models in Medical Education: Opportunities, Challenges, and Future Directions. *JMIR Med Educ* 2023; **9**: e48291.

13 Minssen T, Vayena E, Cohen IG. The Challenges for Regulating Medical Use of ChatGPT and Other Large Language Models. *JAMA* 2023; **330**: 315–6.

14 ChatGPT banned in Italy over privacy concerns. BBC News. 2023; published online March 31. https://www.bbc.com/news/technology-65139406 (accessed Oct 24, 2023).

15 ChatGPT accessible again in Italy. BBC News. 2023; published online April 28. https://www.bbc.com/news/technology-65431914 (accessed Oct 24, 2023).

16 Goodfellow I, Pouget-Abadie J, Mirza M, *et al.* Generative Adversarial Nets. In: Advances in Neural Information Processing Systems. Curran Associates, Inc., 2014. https://papers.nips.cc/paper_files/paper/2014/hash/5ca3e9b122f61f8f06494c97b1afccf3-Abstract.html (accessed Oct 23, 2023).

17 Ahmad W, Ali H, Shah Z, Azmat S. A new generative adversarial network for medical images super resolution. *Sci Rep* 2022; **12**: 9533.

18 Cai Z, Xiong Z, Xu H, Wang P, Li W, Pan Y. Generative Adversarial Networks: A Survey Toward Private and Secure Applications. *ACM Comput Surv* 2021; **54**: 132:1-132:38.

19 Park C, Jeong HK, Henao R, Kheterpal M. Current Landscape of Generative Adversarial Networks for Facial Deidentification in Dermatology: Systematic Review and Evaluation. *JMIR Dermatol* 2022; **5**: e35497.

20 Thambawita V, Hicks SA, Isaksen J, *et al.* DeepSynthBody: the beginning of the end for data deficiency in medicine. In: 2021 International Conference on Applied Artificial Intelligence (ICAPAI). 2021: 1–8.

21 Hernandez M, Epelde G, Alberdi A, Cilla R, Rankin D. Synthetic data generation for tabular health records: A systematic review. *Neurocomputing* 2022; **493**: 28–45.

22 Li J, Cairns BJ, Li J, Zhu T. Generating synthetic mixed-type longitudinal electronic health records for artificial intelligent applications. *npj Digit Med* 2023; **6**: 1–18.

23 Nikolentzos G, Vazirgiannis M, Xypolopoulos C, Lingman M, Brandt EG. Synthetic electronic health records generated with variational graph autoencoders. *npj Digit Med* 2023; **6**: 1–12.

24 Tricco AC, Lillie E, Zarin W, *et al.* PRISMA Extension for Scoping Reviews (PRISMA-ScR): Checklist and Explanation. *Ann Intern Med* 2018; **169**: 467–73.

25 Fournier-Tombs E, McHardy J. A Medical Ethics Framework for Conversational Artificial Intelligence. *J Med Internet Res* 2023; **25**: e43068.

26 Solanki P, Grundy J, Hussain W. Operationalising ethics in artificial intelligence for healthcare: a framework for AI developers. *AI and Ethics* 2023; **3**: 223–40.

27 Paladugu PS, Ong J, Nelson N, *et al.* Generative Adversarial Networks in Medicine: Important Considerations for this Emerging Innovation in Artificial Intelligence. *Ann Biomed Eng* 2023; **51**: 2130–42.

28 Marks M, Haupt CE. AI Chatbots, Health Privacy, and Challenges to HIPAA Compliance. *JAMA* 2023; **330**: 309–10.

29 Kunze KN, Jang SJ, Fullerton MA, Vigdorchik JM, Haddad FS. What's all the chatter about? *Bone Joint J* 2023; **105-B**: 587–9.

30 Hussain F, Ksantini R, Hammad M. A Review of Malicious Altering Healthcare Imagery using Artificial Intelligence. In: 2021 International Conference on Innovation and Intelligence for Informatics, Computing, and Technologies (3ICT). 2021: 646–51.

31 Sun H, Zhu T, Zhang Z, Jin D, Xiong P, Zhou W. Adversarial Attacks Against Deep Generative Models on Data: A Survey. *IEEE Trans on Knowl and Data Eng* 2023; **35**: 3367–88.

32 Yan C, Yan Y, Wan Z, *et al.* A Multifaceted benchmarking of synthetic electronic health record generation models. *Nat Commun* 2022; **13**: 7609.

33 Kuo NI-H, Garcia F, Sönnerborg A, *et al.* Generating synthetic clinical data that capture class imbalanced distributions with generative adversarial networks: Example using antiretroviral therapy for HIV. *Journal of Biomedical Informatics* 2023; **144**: 104436.

34 Singhal K, Azizi S, Tu T, *et al.* Large language models encode clinical knowledge. *Nature* 2023; **620**: 172–80.

35 Dado T, Güçlütürk Y, Ambrogioni L, *et al.* Hyperrealistic neural decoding for reconstructing faces from fMRI activations via the GAN latent space. *Sci Rep* 2022; **12**: 141.

36 Morley J, Machado CCV, Burr C, *et al.* The ethics of AI in health care: A mapping review. *Social Science & Medicine* 2020; **260**: 113172.

37 Ghallab M. Responsible AI: requirements and challenges. *AI Perspectives* 2019; **1**: 3.

38 Acosta JN, Falcone GJ, Rajpurkar P, Topol EJ. Multimodal biomedical AI. *Nat Med* 2022; **28**: 1773–84.

39 Multimodal generative AI search. Google Cloud Blog. https://cloud.google.com/blog/products/ai-machine-learning/multimodal-generative-ai-search (accessed Oct 17, 2023).

40 ChatGPT can now see, hear, and speak. https://openai.com/blog/chatgpt-can-now-see-hear-and-speak (accessed Oct 23, 2023).

41 Bakhshandeh S. Benchmarking medical large language models. *Nat Rev Bioeng* 2023; **1**: 543–543.

42 Tang L, Sun Z, Idnay B, *et al.* Evaluating large language models on medical evidence summarization. *npj Digit Med* 2023; **6**: 1–8.

43 Wornow M, Xu Y, Thapa R, *et al.* The shaky foundations of large language models and foundation models for electronic health records. *npj Digit Med* 2023; **6**: 1–10.

44 Takagi Y, Nishimoto S. High-resolution image reconstruction with latent diffusion models from human brain activity. In: 2023 IEEE/CVF Conference on Computer Vision and Pattern Recognition (CVPR). Los Alamitos, CA, USA: IEEE Computer Society, 2023: 14453–63.

45 Chen Z, Qing J, Xiang T, Yue W, Zhou J. Seeing Beyond the Brain: Conditional Diffusion Model with Sparse Masked Modeling for Vision Decoding. In: 2023 IEEE/CVF Conference on Computer Vision and Pattern Recognition (CVPR). Los Alamitos, CA, USA: IEEE Computer Society, 2023: 22710–20.

46 Tang J, LeBel A, Jain S, Huth AG. Semantic reconstruction of continuous language from non-invasive brain recordings. *Nat Neurosci* 2023; **26**: 858–66.

47 Rainey S, Martin S, Christen A, Mégevand P, Fourneret E. Brain Recording, Mind-Reading, and Neurotechnology: Ethical Issues from Consumer Devices to Brain-Based Speech Decoding. *Sci Eng Ethics* 2020; **26**: 2295–311.

48 Vidal C. Neurotechnologies under the Eye of Bioethics. *eNeuro* 2022; **9**. DOI:10.1523/ENEURO.0072-22.2022.

49 Porsdam Mann S, Earp BD, Nyholm S, *et al.* Generative AI entails a credit–blame asymmetry. *Nat Mach Intell* 2023; **5**: 472–5.

50 Catchpole K, Russ S. The problem with checklists. *BMJ Qual Saf* 2015; **24**: 545–9.

51 Liu M, Ning Y, Teixayavong S, *et al.* A translational perspective towards clinical AI fairness. *npj Digit Med* 2023; **6**: 1–6.

52 Collins GS, Reitsma JB, Altman DG, Moons KGM. Transparent reporting of a multivariable prediction model for individual prognosis or diagnosis (TRIPOD): the TRIPOD statement. *BMJ (Clinical research ed)* 2015; **350**: g7594.

53 Collins GS, Dhiman P, Andaur Navarro CL, *et al.* Protocol for development of a reporting guideline (TRIPOD-AI) and risk of bias tool (PROBAST-AI) for diagnostic and prognostic prediction model studies based on artificial intelligence. *BMJ open* 2021; **11**: e048008.

54 Mongan J, Moy L, Charles E. Kahn J. Checklist for Artificial Intelligence in Medical Imaging (CLAIM): A Guide for Authors and Reviewers. *Radiology: Artificial Intelligence* 2020; published online March 25. DOI:10.1148/ryai.2020200029.

55 Tejani AS, Klontzas ME, Gatti AA, *et al.* Updating the Checklist for Artificial Intelligence in Medical Imaging (CLAIM) for reporting AI research. *Nat Mach Intell* 2023; **5**: 950–1.

56 Caixa F la. Ethical OS Toolkit - a guide to anticipating the future impact of today's technology. https://rri-tools.eu/-/ethical-os-toolkit (accessed Nov 2, 2023).

57 Cacciamani GE, Eppler MB, Ganjavi C, *et al.* Development of the ChatGPT, Generative Artificial Intelligence and Natural Large Language Models for Accountable Reporting and Use (CANGARU) Guidelines. 2023; published online July 18. DOI:10.48550/arXiv.2307.08974.

58 Murphy K, Di Ruggiero E, Upshur R, *et al.* Artificial intelligence for good health: a scoping review of the ethics literature. *BMC Medical Ethics* 2021; **22**: 14.

59 WHO guidance. Ethics and governance of artificial intelligence for health. https://www.who.int/publications-detail-redirect/9789240029200 (accessed Dec 20, 2023).

**Box 1.** Definition of the nine ethical principles in healthcare context.

| Ethical Principles | In-Context Definition(s) |
|---|---|
| Accountability | • The explicit clarifications of to whom and to what extent responsibility and/or legal liability fall.<br>• The mandated and moral duty to establish regulatory mechanisms to prevent potential adverse effects on patients from the use of generative AI. |
| Autonomy | • The preservation and fostering of patients' dignity, rights for self-determination and capacity to make informed decisions.<br>• Provision on understandable information to enable patients to employ according to their values. |
| Equity | • The use of generative AI to promote equity according to some notion of fairness (e.g., equality of opportunity, outcomes, etc.) in health or health resources across diverse groups of patient populations, and to actively prevent and/or remedy systemic, unfavourable outcomes in specific patient population(s).<br>• The equitable access to AI or generative AI technology. |
| Integrity (in medical education and quality of clinical research) | • The commitment to intellectual honesty, and personal responsibility to abide by responsible research conduct, including data integrity, to establish accountability and prevent harm.<br>• The rightful acknowledgement of contributions to and ownership of intellectual work, when generative AI is used in clinical research. |
| Non-maleficence | • The prevention of harm and potential risks to patients associated with generative AI use in healthcare. |
| Privacy | • The protection of patients' information from illegitimate access, and of confidentiality of personal sensitive information. |
| Security | • The protection of health data integrity and safety, through careful assessments of vulnerabilities in data systems and the prevention of data breaches, cyberattacks or other threats. |
| Transparency | • The full-disclosure and thorough documentation of information regarding generative AI development, including its data set and evaluation of performance.<br>• The ability to access and understand the processes underlying models' outputs, especially pertaining to black-box models, in so far as this is possible |
| Trust | • The confidence of users in generative AI and/or its developers, and expectations that the model is competent in performing its pre-specified tasks and behave in ways that serve patients and medical community.<br>• Evidence of performance and its limitations<br>• The willingness to accept and integrate generative AI tools to assist delivery of care or research.<br>• Trustworthy generative AI possesses and exhibition of a range of ethically-reliable properties, including performance robustness, fairness, security, etc. |

**Table 1.** Codes associated with the nine ethical principles defined in Box 1.

| **Ethical Principles** | **Codes** |
|---|---|
| Accountability | Governance, responsibility, accountability, legal, liability |
| Autonomy | Respect, human autonomy, human oversight, informed decisions, informed consent, valid consent |
| Equity | Fair, fairness, bias, disparity, discrimination, justice, equity, equality, inequity |
| Integrity (in medical education and quality of clinical research) | Integrity, plagiarism, data integrity, copyright, ownership, intellectual property |
| Non-maleficence | Safe, safety, harm, harmful, misinformation, truthful, risk, benevolence |
| Privacy | Privacy, private, confidential, confidentiality |
| Security | Security, cybersecurity |
| Transparency | Transparent, transparency, explainable, explainability, explicability, interpretable, interpretability, non-interpretability, black-box, opacity, white-box |
| Trust | Trust, trustworthy, trustworthiness |

**Table 2.** The Transparent Reporting of Ethics for Generative Artificial Intelligence (TREGAI) checklist. Beneficence refers to the putative benefits of AI tools offer and the limits of these benefits. Refer to Box 1 for the in-context definitions of other ethical principles.

| **Ethicists involvement** | | | | |
|---|---|---|---|---|
| **For authors/researchers** | ☐ Ethical discussions disclosed below involved contributions from philosophers and/or ethicists. | | | |
| **For journal/funder** | ☐ Ethical discussions disclosed below were peer-reviewed by philosophers and/or ethicists. | | | |
| **Ethical discussions**<br>**For authors:** indicate the presence of discussion on each ethical principle in the manuscript text (check all that apply), and the position of corresponding text of discussion if Yes is checked. | | | | |
| **Ethical Principles** | **I. Discussed issues arising from generative AI** | **II. Further discussed possible solutions to I** | **III. Discussed use of generative AI to resolve issues** | **IV. Further discussed limitations of III** |
| Accountability | ☐ Yes: ________<br>☐ Not applicable | ☐ Yes: ________<br>☐ Not applicable | ☐ Yes: ________<br>☐ Not applicable | ☐ Yes: ________<br>☐ Not applicable |
| Autonomy | ☐ Yes: ________<br>☐ Not applicable | ☐ Yes: ________<br>☐ Not applicable | ☐ Yes: ________<br>☐ Not applicable | ☐ Yes: ________<br>☐ Not applicable |
| Beneficence | ☐ Yes: ________<br>☐ Not applicable | ☐ Yes: ________<br>☐ Not applicable | ☐ Yes: ________<br>☐ Not applicable | ☐ Yes: ________<br>☐ Not applicable |
| Equity | ☐ Yes: ________<br>☐ Not applicable | ☐ Yes: ________<br>☐ Not applicable | ☐ Yes: ________<br>☐ Not applicable | ☐ Yes: ________<br>☐ Not applicable |
| Integrity | ☐ Yes: ________<br>☐ Not applicable | ☐ Yes: ________<br>☐ Not applicable | ☐ Yes: ________<br>☐ Not applicable | ☐ Yes: ________<br>☐ Not applicable |
| Non-maleficence | ☐ Yes: ________<br>☐ Not applicable | ☐ Yes: ________<br>☐ Not applicable | ☐ Yes: ________<br>☐ Not applicable | ☐ Yes: ________<br>☐ Not applicable |
| Privacy | ☐ Yes: ________<br>☐ Not applicable | ☐ Yes: ________<br>☐ Not applicable | ☐ Yes: ________<br>☐ Not applicable | ☐ Yes: ________<br>☐ Not applicable |
| Security | ☐ Yes: ________ | ☐ Yes: ________ | ☐ Yes: ________ | ☐ Yes: ________ |
| | ☐ Not applicable | ☐ Not applicable | ☐ Not applicable | ☐ Not applicable |
| Transparency | ☐ Yes: _________<br>☐ Not applicable | ☐ Yes: _________<br>☐ Not applicable | ☐ Yes: _________<br>☐ Not applicable | ☐ Yes: _________<br>☐ Not applicable |
| Trust | ☐ Yes: _________<br>☐ Not applicable | ☐ Yes: _________<br>☐ Not applicable | ☐ Yes: _________<br>☐ Not applicable | ☐ Yes: _________<br>☐ Not applicable |
| Others: _______ | ☐ Yes: _________<br>☐ Not applicable | ☐ Yes: _________<br>☐ Not applicable | ☐ Yes: _________<br>☐ Not applicable | ☐ Yes: _________<br>☐ Not applicable |

**Figure 1.** Evidence gap map for ethical issues caused or resolved by generative AI methods by data modality. One article may discuss multiple data modalities and ethical issues caused and/or resolved by generative AI.

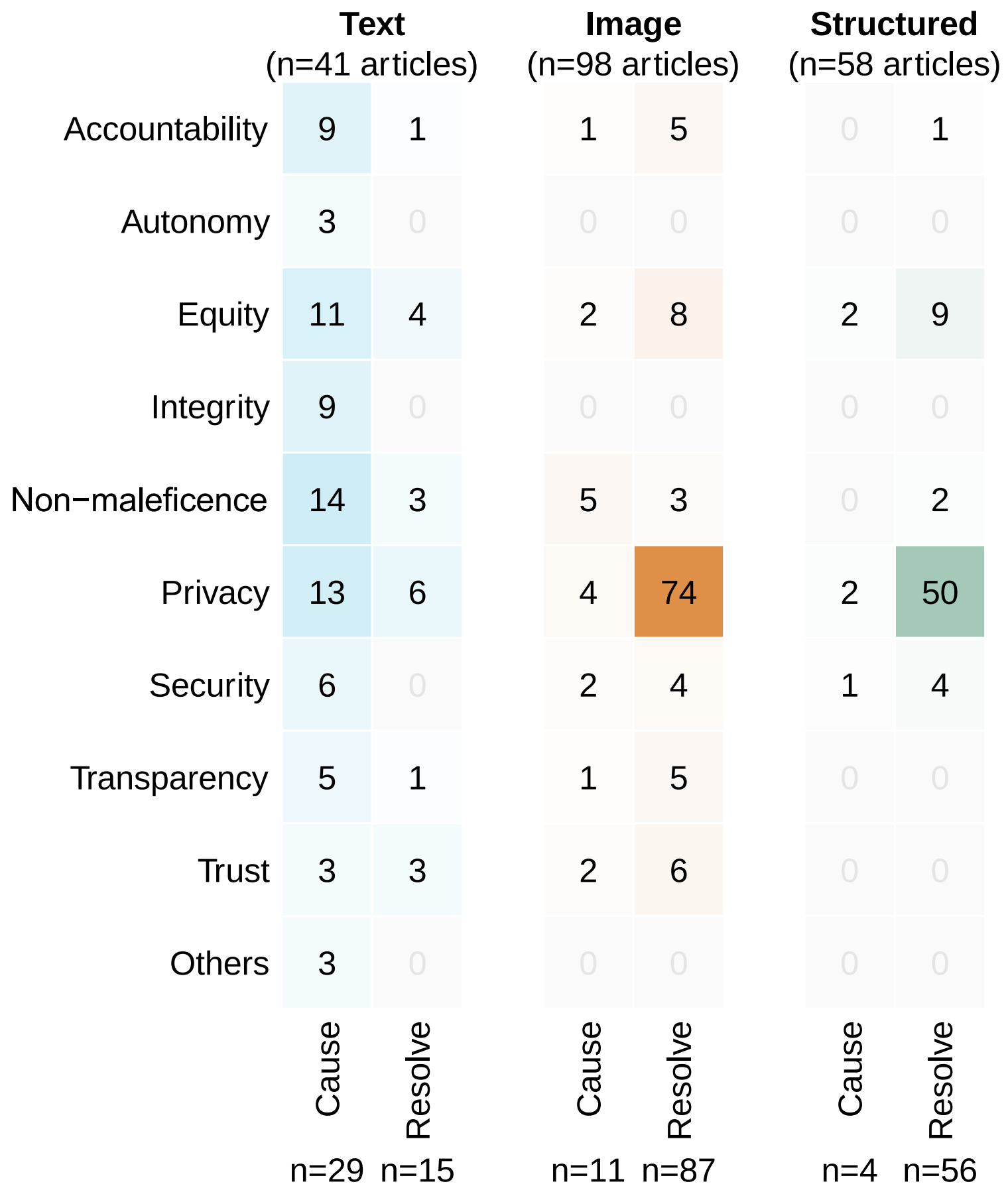